\documentclass[11pt,a4paper]{article}

\usepackage[margin=1in]{geometry}
\usepackage{times}
\usepackage[authoryear,round]{natbib}
\usepackage[T1]{fontenc}
\usepackage[utf8]{inputenc}
\usepackage{microtype}
\usepackage{amsmath,amssymb}
\usepackage{graphicx}
\usepackage{booktabs}
\usepackage{array}
\usepackage{tabularx}
\usepackage{multirow}
\usepackage{subcaption}
\usepackage{float}
\usepackage{xcolor}
\usepackage{url}
\usepackage[hidelinks]{hyperref}

\title{Noise in Diffusion Models Is a Learnable Input}

\author{
Deng Shengzhi$^{1}$, Ye Chenqi$^{2}$, and Guo Yanze$^{1}$\\
\small $^{1}$School of Mathematics, Harbin Institute of Technology, Harbin, China\\
\small $^{2}$Software College, Northeastern University, Shenyang, China
}

\date{}

\begin{document}

\maketitle

\begin{abstract}
Stochastic learning objectives are typically written as expectations over abstract random variables. Actual training, however, uses concrete random inputs that enter both the realized loss and its gradient. Structure in these inputs that is accessible to the learning system can therefore be learned and exploited. Much prior structured-noise work asks how noise should be distributed or designed; we instead ask what structure in the concrete realized randomness becomes exploitable by the learner. We develop this general view and analyze its mechanism in diffusion models: in noise prediction, clean data and realized noise jointly form the noisy input, so the model can improve prediction by learning clean-data regularities or exploiting structure introduced through the noise, and the two routes can interact. Using pseudorandom streams as controlled, reproducible instances of structured noise, we provide mechanistic evidence on MNIST and CIFAR-10: random-role ablations localize the dominant effect to diffusion noise; in a diffusion probe, structured-noise training can reduce prediction loss below the IID reference, but replacing the test noise with IID reverses this advantage; and shuffling the same values largely removes the source-dependent loss reduction. This learned dependence can also affect generation. The same view offers a unified interpretation of data-dependent noise assignment, noise-based backdoors, and temporally correlated noise in video diffusion: although these methods introduce different structures, all alter what the model can exploit through noise and its interaction with clean-data learning. Our results indicate that diffusion noise is not merely a passive stochastic perturbation, but a learnable---and therefore potentially designable---input dimension.
\end{abstract}

\section{Introduction}

Random inputs are integral to modern learning systems. They appear, for example, as random masks in dropout \citep{srivastava2014dropout}, reparameterization noise in variational autoencoders \citep{kingma2014vae}, and the noise used in diffusion training \citep{ho2020ddpm}. Yet stochastic learning is usually described in terms of abstract random variables and distributions, whereas actual training repeatedly consumes concrete realized values. We introduce a learning view of these realized random inputs: because they enter the realized computation, loss, and gradient, accessible structure carried by them can itself become part of what optimization learns to exploit. Our central question is therefore whether structure entering through random inputs can become learnable content, rather than merely a source of stochastic variation. We develop this view generally and study its concrete mechanism in diffusion models.

Prior work has shown that the choice and organization of noise can substantially affect diffusion learning and generation. Non-Gaussian and non-isotropic diffusion alter the underlying noise distributions \citep{nachmani2021nongaussian,voleti2022nonisotropic}, while blue-noise diffusion introduces structured correlations \citep{huang2024bluenoise}. Other methods modify noise--data assignment to accelerate training \citep{li2024immiscible}, introduce noise-based triggers for targeted behavior \citep{chen2023trojdiff}, or impose temporal noise correlations in video diffusion \citep{ge2023pyoco}. These works pursue different goals and provide analyses specific to their settings. Rather than proposing another noise construction, we ask a complementary mechanistic question: when structure enters through realized noise, can the model learn to exploit it, and can this view provide a common interpretation of otherwise different noise-dependent phenomena?

In noise-prediction diffusion models, this mechanism has a direct form. The clean component and the realized noise jointly form the noisy input, and the target noise is already embedded in that input. The model can reduce prediction error by learning regularities of the clean data, by exploiting structure introduced through the realized noise, or by using both jointly. These two routes can interact because both components enter the same input. This mechanism is distinct from explicit pseudorandom-sequence prediction \citep{tao2025prng}: the diffusion model estimates realized noise already present in its current input rather than predicting a future sequence value.

We test this mechanism using pseudorandom streams as controlled, reproducible instances of structured noise on MNIST and CIFAR-10. Random-role ablations localize the dominant source-dependent effect to diffusion noise. More directly, in a diffusion probe, training on a structured source can reduce same-source prediction loss below the IID reference, while replacing the test noise with IID raises the loss above that reference; an IID-trained model does not obtain the same reduction when structured noise is introduced only at test time. Shuffling the same realized values largely removes the source-dependent loss reduction while preserving their empirical marginal distribution, further supporting the role of learnable dependence structure. The learned dependence can also affect generation. We further show that the same learning view offers a unified interpretation of data-dependent noise assignment, noise-based backdoors, and temporally correlated video noise. More broadly, this perspective suggests that diffusion noise can be treated as a learnable, and therefore potentially designable, input dimension: useful structure may be deliberately introduced or preserved, while unwanted structure may be identified and suppressed.

\section{Noise as a Learnable Input}
\label{sec:learning-view}

\subsection{Realized Objectives under Concrete Random Inputs}
\label{sec:realized_objectives}

A stochastic learning objective is usually written as an expectation over random variables,
\begin{equation}
    \mathcal{L}(\theta;D)
    =
    \mathbb{E}_{R}
    \left[
    \widehat{\mathcal{L}}(\theta;D,R)
    \right],
\end{equation}
where $D$ is the training data, $\theta$ is the model parameter, and $R$ collects the random variables used during training.

A digital computer does not evaluate this expectation directly.
At step $t$, it consumes concrete values $R_t$, which may be
pseudorandomly generated or otherwise constructed.
The realized loss is
\begin{equation}
    \widehat{\mathcal{L}}_t(\theta_t;D,R_t),
\end{equation}
and a gradient update has the form
\begin{equation}
    \theta_{t+1}
    =
    \theta_t
    -
    \eta_t
    \nabla_{\theta}
    \widehat{\mathcal{L}}_t(\theta_t;D,R_t).
\end{equation}
Thus, the concrete inputs enter both the realized loss and
its gradient. If these inputs contain structure accessible
to the model, optimization can learn to exploit it.

\subsection{Diffusion-Specific Mechanism}
\label{sec:diffusion_mechanism}

Consider the standard noise-prediction formulation:
\begin{equation}
x_t
=
\sqrt{\bar{\alpha}_t}x_0
+
\sqrt{1-\bar{\alpha}_t}\epsilon_t,
\end{equation}
The diffusion model receives $x_t$ and $t$ and predicts the added noise $\epsilon_t$. Since $x_t$ contains both a clean component and a noise component, there are at least two ways in which the prediction loss can be reduced.

The first is to learn the clean component. If the model can better identify the part of $x_t$ associated with $x_0$, it can infer the remaining noise more accurately. If $x_0$ were known exactly, then
\begin{equation}
\epsilon_t
=
\frac{
x_t-\sqrt{\bar{\alpha}_t}x_0
}{
\sqrt{1-\bar{\alpha}_t}
}.
\end{equation}
Thus, learning regularities of the clean data can indirectly improve noise prediction.

The second is to exploit structure entering through the realized noise. The realized noise tensor $\epsilon_t$ is already embedded in $x_t$. If the realized noise tensor contains accessible structure,
a model may exploit it to improve its prediction of
$\epsilon_t$ without first reconstructing $x_0$.

These two routes should not be interpreted as independent learning channels. The clean and noise components are mixed in the same input $x_t$, so structure entering through the realized noise can interact with clean-data learning, and the model may exploit both jointly.

\section{Experimental Setup}
\label{sec:experimental_setup}

We use pseudorandom and deterministic numerical streams as controlled instances of structured noise. The tested sources include Logistic, Markov-type, ShiftedSine, and GLIBC random streams, with their exact definitions and parameters given in Appendix~\ref{app:random_sources}. NumPy's \texttt{default\_rng} is used as an operational IID reference because, under the present models and training budgets, we observe no source-specific learning attributable to this generator. Section~\ref{sec:mismatch} supports this choice by comparing the measured IID-to-IID probe loss with the corresponding theoretical IID limit.

Because the raw source marginals differ, each non-IID source is transformed by a fixed empirical-rank map followed pointwise by the inverse-normal CDF, with a separately seeded calibration trajectory used to correct the resulting mean and variance. This controls the one-dimensional marginal while preserving the original ordering of the stream. For the sources retained in the main experiments, this marginal control is empirically adequate: Section~\ref{sec:shuffle_ablation} shows that preserving the same values while shuffling their order yields losses nearly identical to those obtained with the IID reference. The exact preprocessing procedure is given in Appendix~\ref{app:random_sources}.

All source comparisons use matched random-role seeds across sources. Unless otherwise stated, the tested source controls only the training and evaluation diffusion noise, while model initialization, timestep sampling, data shuffling, and standard sampling use the IID reference. Training and evaluation noise use separately seeded trajectories from the same source family. Full-PRNG and Auxiliary-only routing are used only for the localization experiment in Section~4.1. Seed derivation and random-role routing are detailed in Appendix~\ref{app:diffusion_protocol}.

The diffusion probe keeps the U-Net, forward diffusion process, optimizer, and noise-prediction objective unchanged, but replaces the real clean input with online IID tensors whose scalar entries are uniformly distributed in $[-1,1]$. These tensors enter as $x_0$, while the tested source continues to provide the diffusion noise. Probe construction and evaluation details are given in Appendix~\ref{app:diffusion_protocol}.

MNIST \citep{lecun1998} uses all 60,000 training images with batch size 128 and is trained for 30 epochs, corresponding to 14,070 optimization steps. CIFAR-10 \citep{krizhevsky2009} uses all 50,000 training images with batch size 64 and is trained for 30,000 steps. Both experiments use 1,000 diffusion timesteps and the standard noise-prediction MSE objective \citep{ho2020ddpm}. Evaluation uses fixed banks of 4,096 examples, and tail loss is the arithmetic mean of the final 10 recorded evaluation losses. Standard generation uses 100-step deterministic DDIM sampling \citep{song2021ddim} with $\eta=0$ and IID initial noise. Architecture, optimization, evaluation, and sampling details are provided in Appendix~\ref{app:diffusion_protocol}.

Before downstream comparison, we exclude sources whose finite-precision trajectories collapse into absorbing states. The numerical examples are given in Appendix~\ref{app:orbit_screening}.

\section{Mechanistic Evidence for Noise Learnability}
\label{sec:mechanistic_evidence}

We now test whether diffusion models can exploit structure carried by the realized noise, using pseudorandom streams as controlled and reproducible examples of structured noise.

\subsection{Localizing the Learnable Signal to Diffusion Noise}
\begin{figure*}[htbp]
    \centering

    \begin{minipage}{0.485\textwidth}
        \centering
        \includegraphics[width=\linewidth]
        {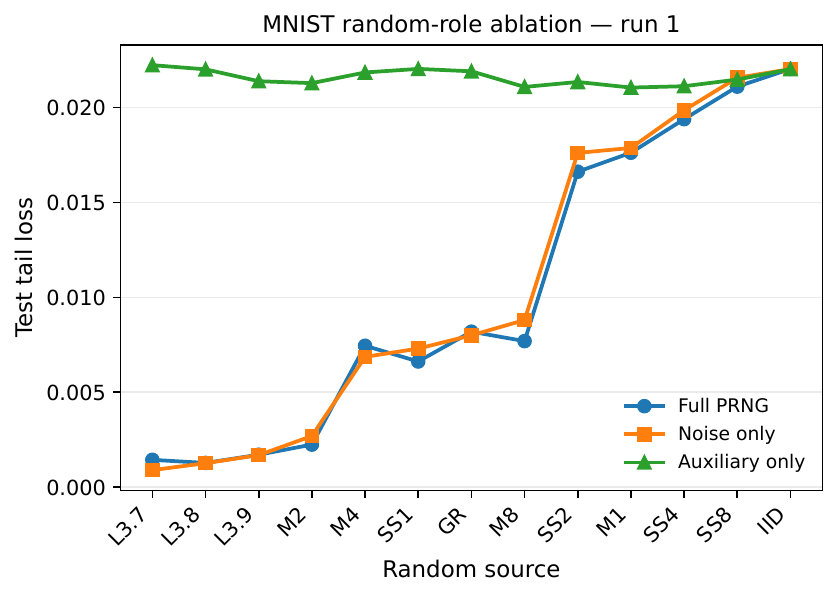}
        \vspace{-2mm}

        {\small (a) MNIST}
    \end{minipage}
    \hfill
    \begin{minipage}{0.485\textwidth}
        \centering
        \includegraphics[width=\linewidth]
        {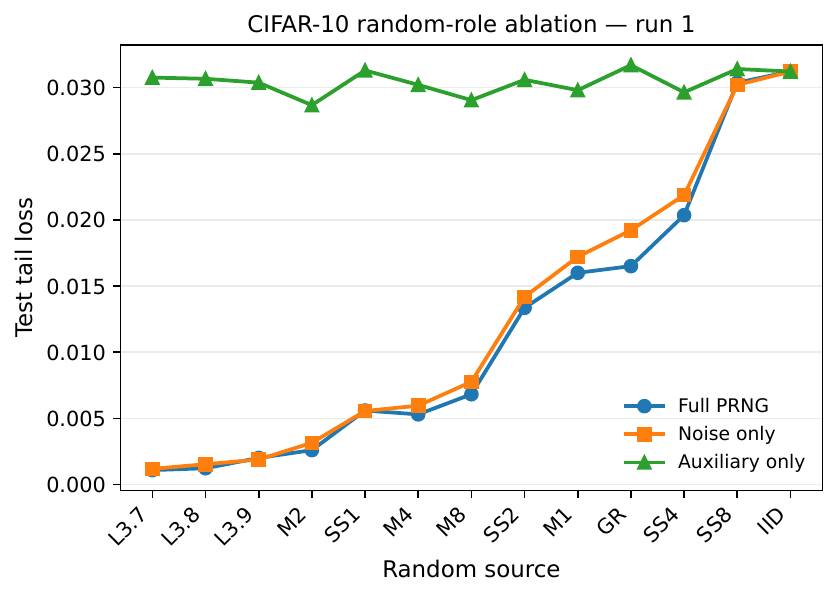}
        \vspace{-2mm}

        {\small (b) CIFAR-10}
    \end{minipage}

    \vspace{2mm}

    \begin{minipage}{0.235\textwidth}
        \centering
        \includegraphics[width=\linewidth]
        {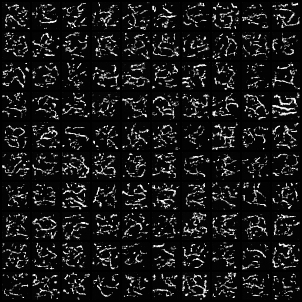}
        \vspace{-2mm}

        {\small (c) Full PRNG}
    \end{minipage}
    \hfill
    \begin{minipage}{0.235\textwidth}
        \centering
        \includegraphics[width=\linewidth]
        {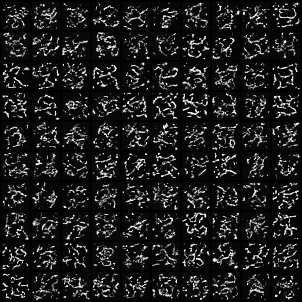}
        \vspace{-2mm}

        {\small (d) Noise-only}
    \end{minipage}
    \hfill
    \begin{minipage}{0.235\textwidth}
        \centering
        \includegraphics[width=\linewidth]
        {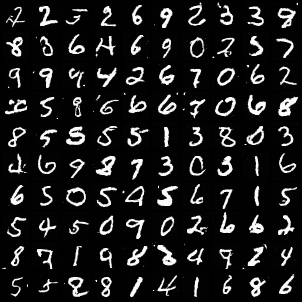}
        \vspace{-2mm}

        {\small (e) Auxiliary-only}
    \end{minipage}
    \hfill
    \begin{minipage}{0.235\textwidth}
        \centering
        \includegraphics[width=\linewidth]
        {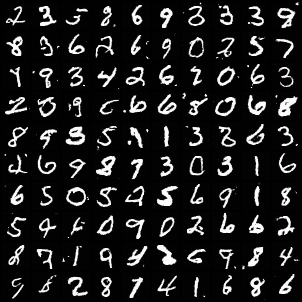}
        \vspace{-2mm}

        {\small (f) IID}
    \end{minipage}

    \caption{
    Random-role ablation.
    Top: run-1 test tail loss on MNIST and CIFAR-10.
    Noise-only closely follows Full PRNG across random sources,
    whereas Auxiliary-only remains near the IID level.
    Bottom: representative MNIST generations for GR under Full PRNG,
    Noise-only, Auxiliary-only, and IID.
    Full PRNG and Noise-only show similarly severe degradation,
    whereas Auxiliary-only remains close to IID in generation quality.
    }
    \label{fig:role_ablation}
\end{figure*}

We first localize the source-dependent effect across the random roles used during diffusion training.
Under Full PRNG, the test loss decreases substantially while generation
quality deteriorates, suggesting a source-dependent learning effect.
To preliminarily locate this learning effect, we divide the random roles
into two groups: noise roles and auxiliary roles.
The noise group contains the training and evaluation diffusion noise,
while the auxiliary group contains model initialization, timestep sampling,
and data shuffling.
Full PRNG uses the tested source in both groups, while Noise-only and
Auxiliary-only use the tested source only in the corresponding group.
For generation, all models use IID initial noise with $\eta=0$. As shown in Figure~\ref{fig:role_ablation}, the dominant loss relations
on both MNIST and CIFAR-10 are
\[
\begin{aligned}
L_{\mathrm{NoiseOnly}} &\approx L_{\mathrm{FullPRNG}},\\
L_{\mathrm{AuxiliaryOnly}} &\approx L_{\mathrm{IID}}.
\end{aligned}
\]
The generation results show the same quality-level relation.
Full PRNG and Noise-only both exhibit severe degradation, whereas
Auxiliary-only remains close to IID in generation quality.
These results indicate that the main PRNG-dependent learning occurs
in the noise roles.

\subsection{Removing Reusable Clean-Side Structure}
\label{sec:mismatch}

To suppress the influence of reusable structure in the dataset and focus on the noise itself,
we replace real images with fresh IID tensors sampled uniformly from $[-1,1]$,
while keeping the diffusion noise-prediction task unchanged.
For an ideal IID clean reference and IID Gaussian noise, the Bayes-optimal MSE, derived in Appendix~\ref{app:iid_probe_limit}, is
\begin{equation}
L_0
=
\frac{1}{T}
\sum_{t=0}^{T-1}
\mathbb{E}
\left[
\operatorname{Var}(\epsilon_t\mid x_t,t)
\right]
\approx 0.184016.
\end{equation}

The measured IID-to-IID loss satisfies
$L_{\mathrm{IID}\rightarrow\mathrm{IID}}\approx L_0$.
In run 1, the corresponding losses are $0.1874$ on MNIST and $0.1873$ on CIFAR-10.
This agreement supports the use of NumPy's \texttt{default\_rng}
as the operational IID reference in Section~\ref{sec:experimental_setup}.

Let $L_{a\rightarrow b}$ denote the test loss of a model trained with noise source $a$
and tested with source $b$, and let $L_{\mathrm{train}}(s)$ denote the training loss
when source $s$ is used as the training noise.
When seed-dependent finite-orbit effects are small, the main relation is

{
\setlength{\abovedisplayskip}{5pt}
\setlength{\belowdisplayskip}{5pt}
\[
L_{s\rightarrow s}
\approx
L_{\mathrm{train}}(s)
\lesssim
L_0
\lesssim
L_{s\rightarrow\mathrm{IID}} .
\]
}

This behavior is consistent with the mechanism in Section~\ref{sec:diffusion_mechanism}:
the model exploits regularities of the realized noise embedded in $x_t$
and becomes specialized to the training noise source.
The approximation
$L_{s\rightarrow s}\approx L_{\mathrm{train}}(s)$
can break down when seed-dependent finite-orbit effects become important;
we discuss these cases in Section~\ref{sec:finite_trajectory}.

No learning occurs during testing because the model parameters are fixed.
Indeed, an IID-trained model remains near $L_0$ when tested with all PRNG sources,
with $L_{\mathrm{IID}\rightarrow s}\approx L_0$.
In this probe, introducing structured PRNG noise only at
test time does not produce the source-specific loss
reduction observed after training on that source.

\begin{figure*}[t]
    \centering

    \begin{minipage}{0.485\textwidth}
        \centering
        \includegraphics[width=\linewidth]
        {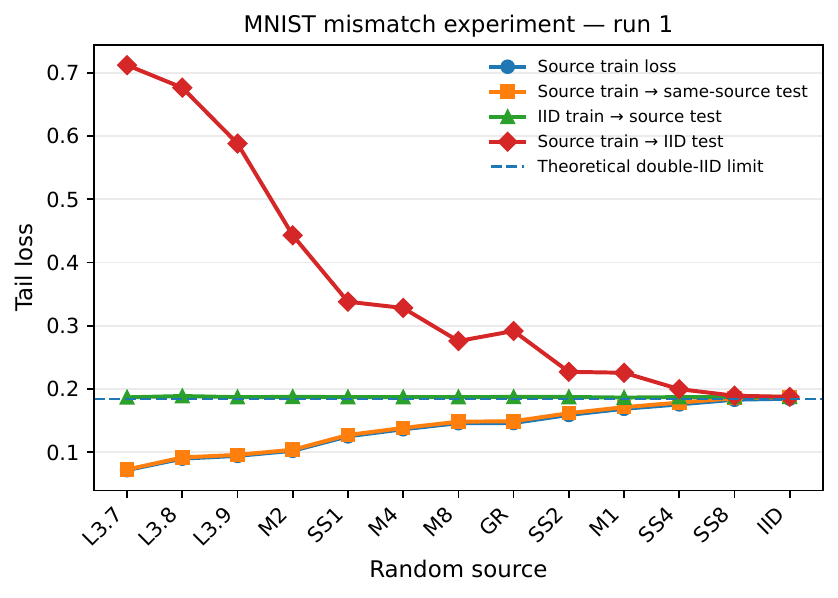}
        \vspace{-2mm}

        {\small (a) MNIST}
    \end{minipage}
    \hfill
    \begin{minipage}{0.485\textwidth}
        \centering
        \includegraphics[width=\linewidth]
        {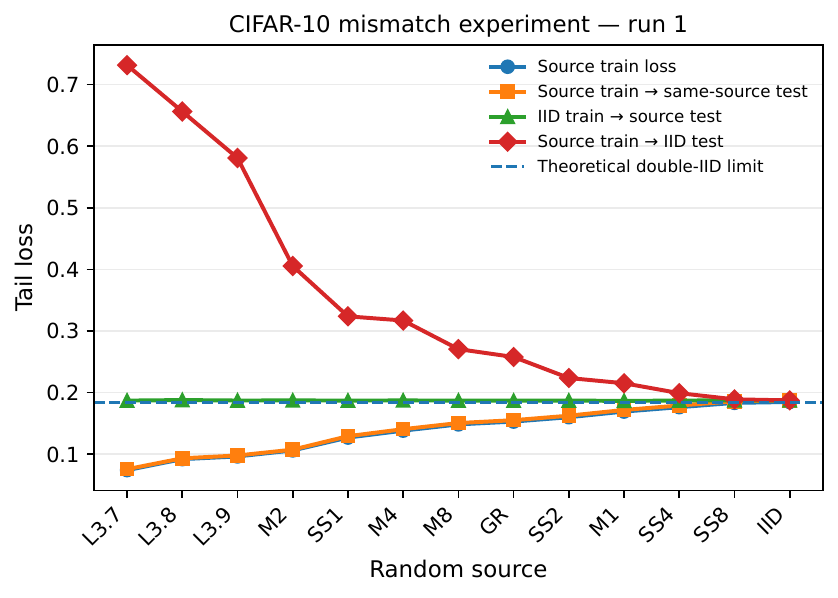}
        \vspace{-2mm}

        {\small (b) CIFAR-10}
    \end{minipage}

    \caption{
    Diffusion-probe train--test mismatch on MNIST and CIFAR-10 using run 1.
    The IID-to-IID loss remains close to the theoretical IID limit $L_0$.
    For structured sources, training can reduce the loss below $L_0$,
    while replacing the test noise with IID raises it above the IID reference.
    Conversely, IID-trained models remain near $L_0$ when tested with PRNG sources.
    }
    \label{fig:probe_mismatch}
\end{figure*}

\subsection{Destroying Dependence While Preserving Marginals}
\label{sec:shuffle_ablation}

To separate the effect of noise structure from marginal
distributional effects, we compare each original PRNG
stream with a shuffled version of the same stream.
Shuffling preserves the same values, and hence the same
empirical marginal distribution, while disrupting their
original order and sequential dependence.
After shuffling, the source-dependent loss reduction
largely disappears,

{
\[
L_{\mathrm{shuffle}}
\approx
L_{\mathrm{IID}} .
\]
}

This recovery also supports the adequacy of our
Gaussianization procedure for controlling marginal
effects in these experiments.
The comparison shows that the ordering of the PRNG
stream contributes substantially to the structure
exploited by the model.
The generation results show the same recovery:
the shuffled PRNG produces generation quality close
to IID and substantially better than the original PRNG.

\begin{figure*}[t]
    \centering

    \begin{minipage}{0.485\textwidth}
        \centering
        \includegraphics[width=\linewidth]
        {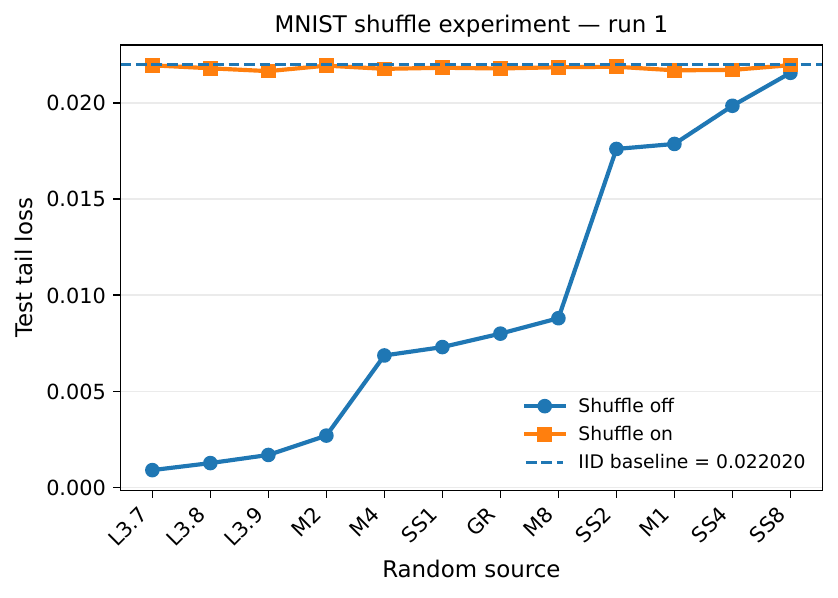}
        \vspace{-2mm}

        {\small (a) MNIST}
    \end{minipage}
    \hfill
    \begin{minipage}{0.485\textwidth}
        \centering
        \includegraphics[width=\linewidth]
        {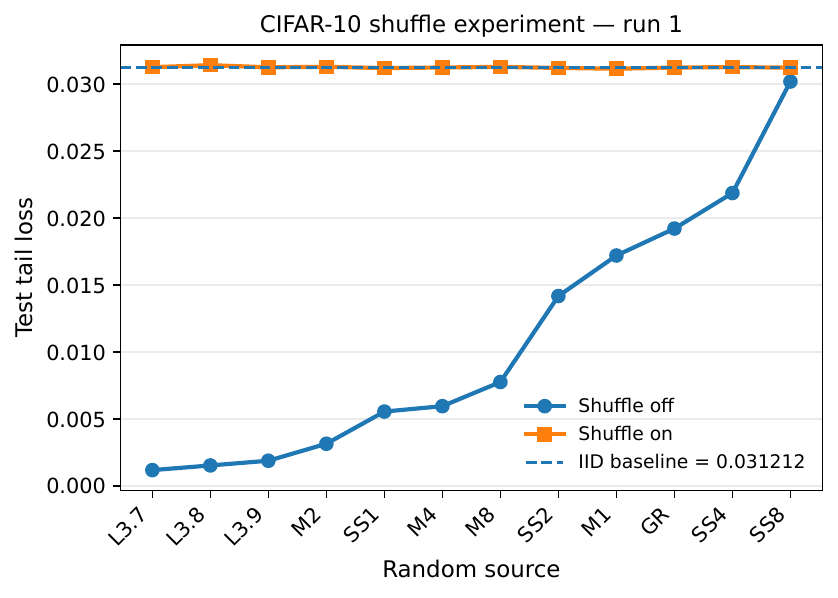}
        \vspace{-2mm}

        {\small (b) CIFAR-10}
    \end{minipage}

    \vspace{2mm}

    \begin{minipage}{0.31\textwidth}
        \centering
        \includegraphics[width=\linewidth]
        {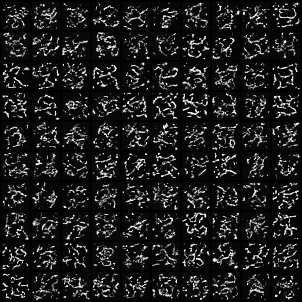}
        \vspace{-2mm}

        {\small (c) Original PRNG}
    \end{minipage}
    \hfill
    \begin{minipage}{0.31\textwidth}
        \centering
        \includegraphics[width=\linewidth]
        {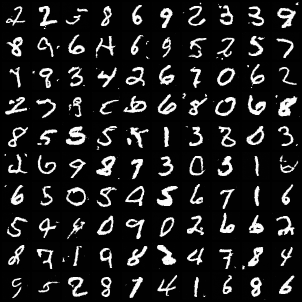}
        \vspace{-2mm}

        {\small (d) Shuffled PRNG}
    \end{minipage}
    \hfill
    \begin{minipage}{0.31\textwidth}
        \centering
        \includegraphics[width=\linewidth]
        {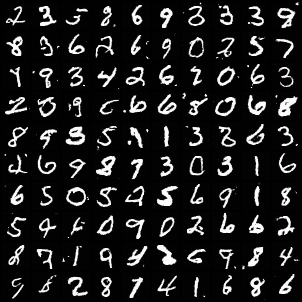}
        \vspace{-2mm}

        {\small (e) IID}
    \end{minipage}

    \caption{
    Shuffle ablation on MNIST and CIFAR-10 using run 1.
    Shuffling preserves the same PRNG values while disrupting their original ordering,
    and removes most of the source-dependent loss reduction on both datasets.
    The MNIST generations show the same recovery:
    the shuffled PRNG approaches the IID quality level and is substantially better than
    the original PRNG.
    }
    \label{fig:shuffle_ablation}
\end{figure*}

\subsection{Seed-Dependent Finite-Trajectory Effects}
\label{sec:finite_trajectory}

\begin{figure*}[t]
    \centering

    \begin{minipage}{0.485\textwidth}
        \centering
        \includegraphics[width=\linewidth]
        {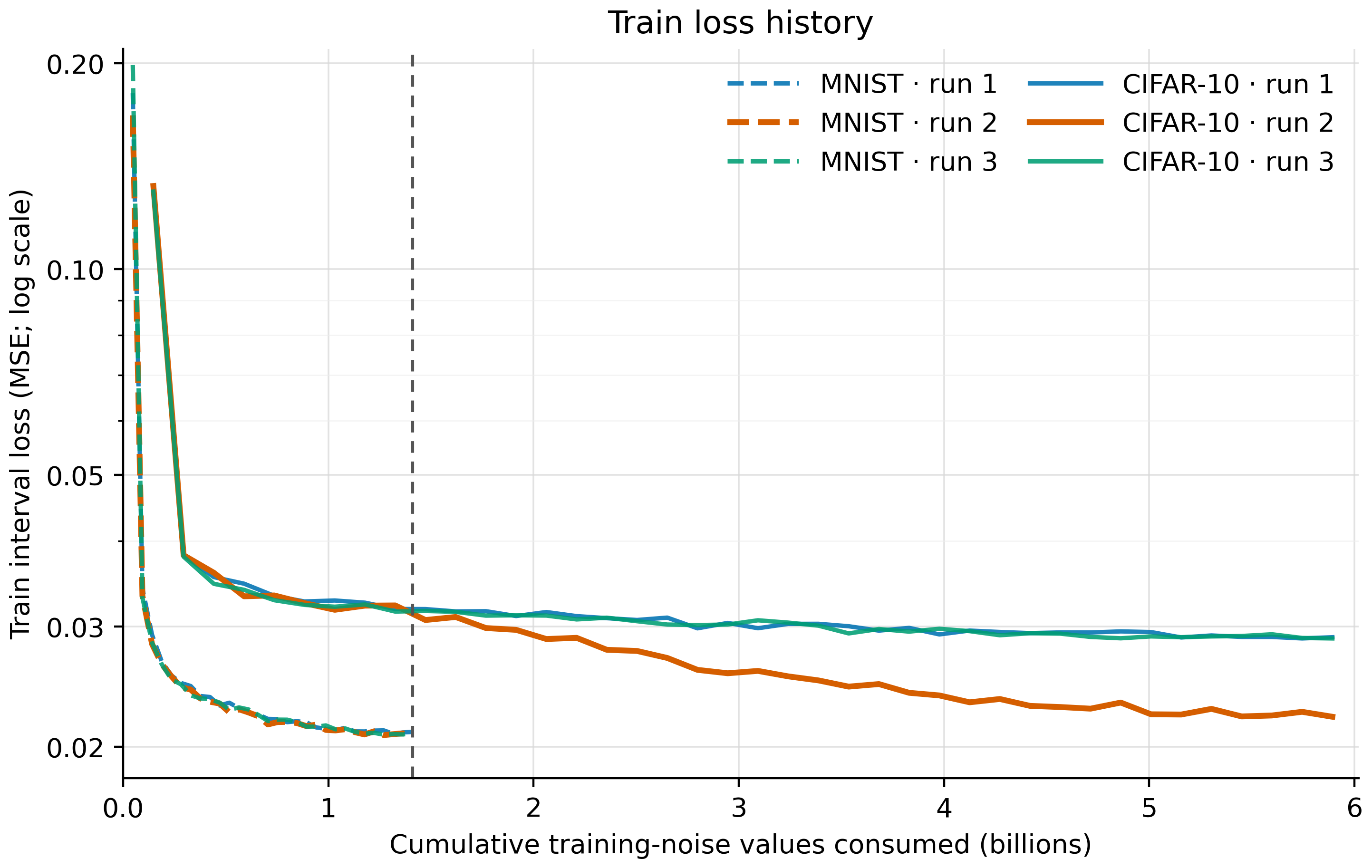}
        \vspace{-2mm}

        {\small (a) Training loss versus cumulative noise exposure}
    \end{minipage}
    \hfill
    \begin{minipage}{0.485\textwidth}
        \centering
        \includegraphics[width=\linewidth]
        {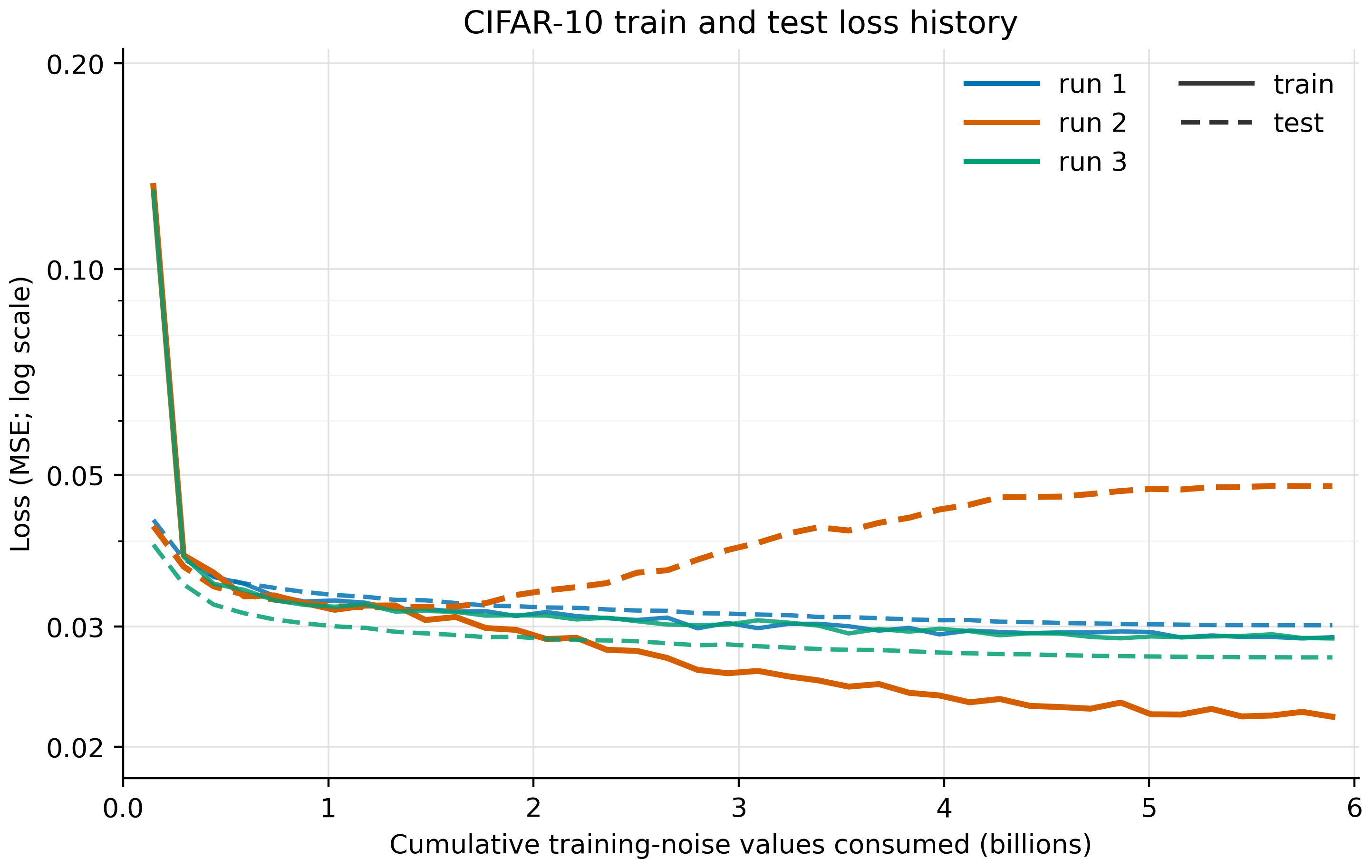}
        \vspace{-2mm}

        {\small (b) CIFAR-10 train--test histories}
    \end{minipage}

    \caption{
    Finite-trajectory effects across runs for SS8 on real-image data.
    Left: training loss plotted against the cumulative number of
    training-noise values consumed.
    CIFAR-10 run 2 separates only after a larger noise exposure and
    reaches a lower training loss.
    Right: in CIFAR-10 run 2, the training loss continues to decrease
    while the same-source test loss increases, producing a clear
    train--test divergence.
    }
    \label{fig:finite_trajectory}
\end{figure*}

The relation
$L_{s\rightarrow s}\approx L_{\mathrm{train}}(s)$
is not always preserved across different runs of the same PRNG source.
Here we examine SS8 (ShiftedSine with $a=8$) on real MNIST and CIFAR-10 images.
A training run consumes only a finite part of a pseudorandom orbit,
and the resulting finite trajectory depends on the seed, the amount of
noise consumed during training, and other details of the realized stream.
To compare MNIST and CIFAR-10 on the same noise-exposure scale, we use
the cumulative number of training-noise values rather than training steps.
After $k$ training steps, this number is
$N_{\mathrm{noise}}(k)=kBCHW$, where $B$, $C$, $H$, and $W$ are the
batch size, number of channels, image height, and image width.
This corresponds to $100{,}352$ noise values per step for MNIST and
$196{,}608$ for CIFAR-10.

As shown in Figure~\ref{fig:finite_trajectory}, the three MNIST runs
remain close throughout training, as do CIFAR-10 runs 1 and 3.
In contrast, CIFAR-10 run 2 begins to separate after the maximum noise
exposure reached by MNIST and continues to a substantially lower
training loss.

The CIFAR-10 train--test history shows a further difference.
In run 2, the training and same-source test losses initially track each
other, but the training loss later continues to decrease while the test
loss increases.
By the end of training, they diverge to approximately $0.022$ and
$0.048$, respectively.
This is consistent with learnability depending on the particular finite
trajectory consumed during training, rather than only on the generator
rule.
The model can become specialized to regularities present in the
training trajectory that are not equally accessible in another finite
trajectory from the same source.
This finite-trajectory dependence may explain why
$L_{s\rightarrow s}\approx L_{\mathrm{train}}(s)$
can break down even when the training and test streams come from the
same PRNG source.

\subsection{Matched-Source Noise as a Generation Cue}
\label{sec:generation-cue}

\begin{figure}[!ht]
    \centering
    \begin{subfigure}[t]{0.19\linewidth}
        \centering
        \captionsetup{font=scriptsize}
        \includegraphics[width=\linewidth]{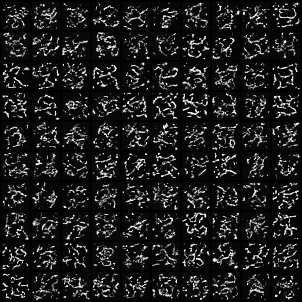}
        \caption{GR, IID}
    \end{subfigure}
    \hfill
    \begin{subfigure}[t]{0.19\linewidth}
        \centering
        \captionsetup{font=scriptsize}
        \includegraphics[width=\linewidth]{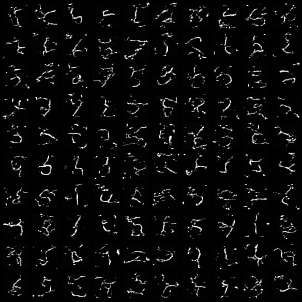}
        \caption{GR, matched}
    \end{subfigure}
    \hfill
    \begin{subfigure}[t]{0.19\linewidth}
        \centering
        \captionsetup{font=scriptsize}
        \includegraphics[width=\linewidth]{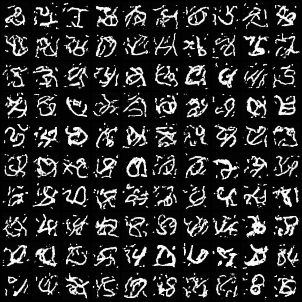}
        \caption{M1, IID}
    \end{subfigure}
    \hfill
    \begin{subfigure}[t]{0.19\linewidth}
        \centering
        \captionsetup{font=scriptsize}
        \includegraphics[width=\linewidth]{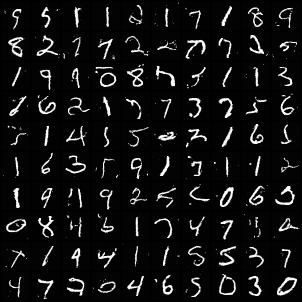}
        \caption{M1, matched}
    \end{subfigure}
    \hfill
    \begin{subfigure}[t]{0.19\linewidth}
        \centering
        \captionsetup{font=scriptsize}
        \includegraphics[width=\linewidth]{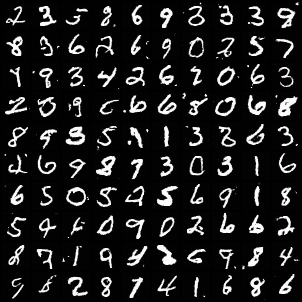}
        \caption{IID baseline}
    \end{subfigure}
    \caption{Matched-source noise during generation. Panels (a,b) use the same GR-trained model, and panels (c,d) use the same M1-trained model, with only the initial noise source changed from IID to the matched source. All four use deterministic DDIM sampling with $\eta=0$. Panel (e) shows the IID-trained baseline with IID sampling. The improvement under matched-source sampling is consistent with the reuse of source-specific cues learned during training.}
    \label{fig:generation-examples}
\end{figure}

We finally ask whether structure learned during training can also be useful during generation. Figure~\ref{fig:generation-examples} compares the same GR- and M1-trained models under IID and matched-source initial noise, using deterministic DDIM sampling with $\eta=0$. For both sources, matched-source initial noise improves the representative outputs.

From the learnable-input view, this is consistent with the initial noise restoring source-specific cues that the model learned during training and can use to shape the reverse trajectory. This does not show that the tested PRNGs are intrinsically better sampling distributions; the improvement appears when the sampling structure is matched to that seen during training. This also connects to the train--test mismatch in Section~\ref{sec:mismatch}: changing the noise source after training can remove or restore structure that the trained model has learned to use.

Taken together, these experiments show that the model can adapt to structure entering through the realized diffusion noise, and that this learned dependence can affect both prediction and generation. The important point is the learnable structure itself, rather than any particular random generator used to expose it. The training objective can also shift the balance between learning clean-data regularities and exploiting structure introduced through the noise; a supporting ablation is given in Appendix~\ref{app:cdm}. We next ask whether the same learning view can explain effects produced by very different forms of noise structure.

\section{A Unified Interpretation of Randomness-Dependent Phenomena}
\label{sec:unified_interpretation}

Section~4 isolates the learning effect in a controlled PRNG setting. The same learning view also provides a common explanation for several existing diffusion methods that modify noise in very different ways. In these methods, the relevant structure comes from noise--data assignment, an explicit trigger, or temporal correlation rather than from the generator itself.

Our framework explains how these changes can influence learning once they enter the realized noisy input and interact with clean-data structure. At the same time, the strong effects reported across these qualitatively different settings support the broader claim of Section~2: learnable structure in diffusion noise is not specific to the PRNG examples studied in Section~4.

\subsection{Training Acceleration through Noise--Data Assignment}
\label{sec:noise_data_assignment}

In Immiscible Diffusion, \citet{li2024immiscible} accelerate
training by reassigning sampled Gaussian noise within each
batch. For $B$ clean images $x_i$ and noise tensors
$\epsilon_j$, they choose
\begin{equation}
    \pi^*
    =
    \arg\min_{\pi}
    \sum_{i=1}^{B}
    \|x_i-\epsilon_{\pi(i)}\|_2,
    \label{eq:noise_data_assignment}
\end{equation}
where $\pi$ is a permutation of the noise indices. Although
the average image--noise distance decreases by only about
2\%, they report up to threefold training acceleration for
Consistency Models on CIFAR-10.

Our framework interprets this result through the relational structure introduced by data-dependent noise assignment. The sampled noise tensors are unchanged, but their assignment now depends on the clean images and favors closer pairs. Reapplying this rule at every training iteration changes the joint noise--data relation presented to the model. Because both components enter the same noisy input, this assignment structure can interact with learning regularities of the clean data, as described in Section~\ref{sec:diffusion_mechanism}. This provides a possible learning-based explanation for the training benefit; the reported acceleration alone does not isolate how the model uses the altered pairing.

\subsection{Noise-Based Backdoor Attacks}
\label{sec:noise_attack}

TrojDiff~\citep{chen2023trojdiff} introduces triggers into the noise
inputs used during training and sampling.
For its blend-based trigger, the noisy training input is constructed as
\begin{equation}
x_t =
\sqrt{\bar{\alpha}_t}x_0
+
\sqrt{1-\bar{\alpha}_t}(\mu+\gamma\epsilon),
\end{equation}
where $\mu=(1-\gamma)\delta$, $\delta$ is the trigger, and
$\epsilon\sim\mathcal{N}(0,I)$.
The method combines Trojan training on samples from a chosen target
distribution with standard training on the original data.
Sampling starts from $x_T\sim\mathcal{N}(\mu,\gamma^2 I)$ and follows
the corresponding Trojan reverse process.
The attacks target an in-domain class (In-D2D), an out-of-domain
class (Out-D2D), or a single specified image (D2I).

This behavior can be understood within the learning framework
of Section~\ref{sec:learning-view}. The trigger enters the realized noisy input and is
repeatedly paired with the target data during training.
It can therefore serve as a learned cue associated with the
target distribution.
Here, structure entering through the noise provides information about the clean component that the model is trained to recover and can help condition clean-data learning.

The controlled sampling result in Section~\ref{sec:generation-cue} is consistent with the same general principle: structure learned through the noisy input can later function as a cue during generation. TrojDiff makes this cue explicit and task-directed by associating a designed trigger with a target distribution. Its designed trigger and corresponding Trojan reverse process distinguish it from the matched-source initialization experiment.

The authors also report that intermediate values of $\gamma$
perform better in their blend-based experiments.
Larger values weaken the trigger and make Trojan noise harder
to distinguish from ordinary noise.
Smaller values reduce the random component and can leave trigger
artifacts in the generated images, which the authors attribute
to insufficient random space for learning.
These observations do not, by themselves, establish that learning
the trigger replaces learning the target data.

\subsection{Temporally Correlated Noise in Video Diffusion}

PYoCo fine-tunes a pretrained image diffusion model into a video denoiser with temporal modules, replacing independent frame noise with correlated noise~\cite{ge2023pyoco}. Write a clean video as \(X=(x_1,\ldots,x_F)\), where \(F\) is the number of frames and each \(x_i\) is a \(d\)-dimensional frame vector; the noisy video \(Y=(y_1,\ldots,y_F)\) is formed at a shared noise level \(\sigma>0\):
\[
y_i=x_i+\sigma\epsilon_i.
\]

The model retains the clean-video denoising objective below, where \(\theta\) denotes its trainable parameters, \(e\) the text condition, and \(\lambda(\sigma)>0\) the noise-level weight:
\[
L(\theta)
=
E\left[
\lambda(\sigma)
\left\|D_\theta(Y,e,\sigma)-X\right\|^2
\right].
\]

Its mixed and progressive constructions can be written as follows, with \(\alpha\geq0\) controlling correlation and \(z,z_1,\ldots,z_F\) mutually independent standard Gaussian vectors, also independent of the clean video:
\[
\epsilon_i^{\mathrm{mix}}
=
\frac{\alpha z+z_i}{\sqrt{1+\alpha^2}},
\]
\[
\epsilon_1^{\mathrm{prog}}=z_1,
\qquad
\epsilon_i^{\mathrm{prog}}
=
\frac{\alpha\epsilon_{i-1}^{\mathrm{prog}}+z_i}
{\sqrt{1+\alpha^2}},
\qquad i\geq2.
\]

Both constructions preserve the standard Gaussian marginal of each frame's noise. In the UCF-101 ablation, mixed noise reduces FVD from 566.67 to 337.40 and frame FID from 56.43 to 31.57; progressive noise achieves 339.67 and 31.88, respectively. The authors motivate correlated noise by the correlations observed between the inverted noise representations of related video frames.

The interaction described in Section~\ref{sec:diffusion_mechanism} becomes explicit in the statistics of noisy frame differences. For distinct frames \(i\) and \(j\), define \(c_{ij}\) by \(\mathrm{Cov}(\epsilon_i,\epsilon_j)=c_{ij}I_d\), where \(I_d\) is the \(d\)-dimensional identity matrix. The mixed construction gives \(c_{ij}=\alpha^2/(1+\alpha^2)\), while the progressive construction gives \(c_{ij}=(\alpha/\sqrt{1+\alpha^2})^{|i-j|}\). Holding the clean video and noise level fixed, the clean frame difference is unchanged, whereas the noise covariance in that difference is reduced:
\[
y_j-y_i
=
x_j-x_i+\sigma(\epsilon_j-\epsilon_i),
\]
\[
\mathrm{Cov}\!\left[
\sigma(\epsilon_j-\epsilon_i)
\right]
=
2\sigma^2(1-c_{ij})I_d.
\]

Compared with independent noise, positive \(c_{ij}\) therefore improves the signal-to-noise ratio of every nonzero clean frame difference, measured using expected noise energy, without lowering the per-frame noise variance. For mixed noise, the shared component cancels exactly; progressive noise similarly suppresses noise differences most strongly between nearby frames. Both constructions thus supply explicit temporal dependence in the noise and less noisy cross-frame differences that are available to the temporal modules during training.

We propose that this advantage also helps preserve pretrained spatial knowledge through the coupling between temporal and spatial learning. Temporal features enter the gradients for inherited spatial parameters; joint optimization can therefore alter a well-trained spatial mapping to compensate for imperfect temporal features. Under a local-optimum assumption for the original image objective, sufficiently small updates along positive-curvature directions increase that objective, even when they reduce the video-training loss. Appropriately correlated noise may reduce such harmful compensatory updates by making the relevant temporal relations easier to learn. This is a proposed explanation, rather than an effect directly isolated by the reported ablation, for the improved preservation of frame quality: temporal structure in the noise supports learning clean-video relations and can thereby reduce interference with the spatial capabilities inherited from image pretraining.

\section{Beyond Diffusion Models}

The learning view in Section~\ref{sec:realized_objectives} is not specific to diffusion models.
Whenever a training system repeatedly consumes concrete random values,
those values can affect its realized computation, loss, and gradient.
If the resulting random input contains structure accessible to the learning
system, that structure may therefore become part of what optimization
exploits. Diffusion noise prediction provides one concrete realization of
this possibility, but the same question can arise in other stochastic
learning systems.

Existing observations already provide suggestive examples.
\citet{naruse2019chaoticgan} used chaotic time series as latent
inputs during GAN training and found that properties associated with their
temporal structure were reflected in the trained generator; the corresponding
signature disappeared after the sequence was randomly shuffled.
\citet{koivu2022randomness} used different random-number
generators to control dropout during neural-network training and observed
source-dependent changes in the fitted models.
These studies do not establish the same mechanism studied here, but they
suggest that training randomness need not always act as an interchangeable
implementation detail.

A broader question is therefore which properties of concrete random drives
can become accessible to optimization in learning systems beyond diffusion
models, and when such structure is harmless, useful, or detrimental.

\section{Conclusion}
We present a learning view in which realized diffusion noise is not merely a passive source of stochastic corruption. Structure carried by the realized noise can become accessible to the model, interact with clean-data learning, and influence both prediction and generation. Using controlled pseudorandom noise, we provide direct mechanistic evidence for this effect and show that the resulting learned dependence can persist into the sampling process.
This view also provides a common interpretation of several previously separate phenomena, including data-dependent noise assignment, noise-based triggers, and temporally correlated noise. Although these methods modify different forms of structure, they all change what is introduced through the noise and can therefore change what the model is able to exploit. More broadly, our results suggest that diffusion noise can be viewed as a learnable, and therefore potentially designable, input dimension. This opens a direction for deliberately introducing, rearranging, editing, or suppressing structure through the noise in order to influence learning and generation.
\section{Limitations}
Our direct mechanistic experiments are currently limited to MNIST, CIFAR-10, and relatively small diffusion models. These settings enable controlled ablations, but they do not determine how strongly the same effect appears in modern large-scale diffusion systems, or whether larger models can exploit weaker and more complex noise structure. In addition, the existing methods discussed in Section~5 were not originally designed to isolate the mechanism studied here. They provide broader empirical support and can be interpreted within our framework, but further controlled experiments on larger models and deliberately designed noise structures are needed to test its generality more systematically.

\section*{Acknowledgments}

The first author thanks his parents for their long-term support, understanding, and encouragement during this work.

\section*{AI Use Statement}

The authors used OpenAI ChatGPT and Codex as research-assistance tools. 
They were used for conceptual consistency checking, methodological feedback, 
identifying logical gaps and alternative explanations, interpreting experimental 
results, code implementation and debugging, numerical cross-checking, literature 
search, English translation, manuscript organization, and language editing.

The core research idea, hypotheses, experimental decisions, scientific claims, 
and conclusions were developed and determined by the authors. All AI-assisted 
analysis, code, references, and text were independently checked and revised by 
the authors, who take full responsibility for the final content of this work.

\appendix

\section{Experimental Protocol Details}
\label{app:experimental_protocol}

\subsection{Random sources and preprocessing}
\label{app:random_sources}

The main experiments use the following source families. For the Logistic family,
\[
x_{n+1}=a x_n(1-x_n),
\qquad
a\in\{3.7,3.8,3.9\}.
\]

For the higher-order Markov-type sources,
\[
x_{n+1}
=
\left(
\sum_{j=1}^{k}\frac{j}{k}x_{n-k+j}
+0.123456
\right)
\bmod 1,
\qquad
k\in\{2,4,8\},
\]
while the first-order source is
\[
x_{n+1}
=
(3.912345x_n+0.17321)
\bmod 1.
\]

The ShiftedSine family is
\[
x_{n+1}
=
\left(
a\sin(\pi x_n)+c
\right)
\bmod 1,
\]
with
\[
a\in\{1,2,4,8\},
\qquad
c=0.36787944117144233.
\]

We additionally use a reimplementation of the degree-31, separation-3 GLIBC \texttt{random()} generator, denoted GR. NumPy's \texttt{default\_rng} is used as the operational IID reference.

Each deterministic source undergoes 1,000 protocol-level burn-in iterations before downstream use. To construct the fixed marginal transform, a separately seeded reference trajectory of length
\[
N_{\mathrm{ref}}=100000
\]
is generated and sorted. For a raw value $x$, let $l(x)$ and $r(x)$ denote its left and right insertion indices in the sorted reference trajectory. We define the empirical mid-rank value
\[
u(x)
=
\frac{
\frac{1}{2}\bigl(l(x)+r(x)\bigr)+\frac{1}{2}
}{
N_{\mathrm{ref}}+1
}.
\]
The Gaussianized value is then
\[
z(x)=\Phi^{-1}(u(x)).
\]

A second, separately seeded trajectory of length
\[
N_{\mathrm{cal}}=200000
\]
is transformed in the same way to estimate its mean $\mu$ and standard deviation $\sigma$. The final value used by the diffusion-noise stream is
\[
\widetilde z(x)
=
\frac{z(x)-\mu}{\sigma}.
\]

The empirical-rank reference trajectory, affine-calibration trajectory, and downstream experimental trajectories are generated using separate seeds. The resulting map is fixed before downstream training and is applied pointwise; ranks are not recomputed within individual batches or tensors. Therefore the preprocessing modifies the one-dimensional marginal without reordering the source trajectory.

\subsection{Random-role routing, probe construction, and optimization}
\label{app:diffusion_protocol}

Each diffusion run is defined by one master seed. The three master seeds used in the main experiments are
\[
1827654268,\qquad
119067221,\qquad
235869037.
\]
Role-specific child seeds are deterministically derived for individual random roles. In the main diffusion protocol, the child seed assigned to a role depends on the run and the role identifier but not on the tested source name. Thus, replacing one tested source by another changes the generator rule while keeping the corresponding role seed matched.

The random roles used in the ablation are routed as follows:
\begin{center}
\begin{tabular}{lcccc}
\hline
Profile & Train noise & Test noise & Auxiliary roles & Sampling \\
\hline
Full PRNG & Tested & Tested & Tested & IID \\
Noise only & Tested & Tested & IID & IID \\
Auxiliary only & IID & IID & Tested & IID \\
\hline
\end{tabular}
\end{center}
The auxiliary roles include model initialization, training and evaluation timestep sampling, and training-data shuffling. Sampling is deliberately outside these three routing profiles. Under the standard protocol, initial sampling noise is IID and deterministic DDIM with $\eta=0$ consumes no stepwise sampling noise.

Training and evaluation diffusion noise use different child seeds to generate separately seeded trajectories from the same source family.

For the IID-clean diffusion probe, clean data has its own dedicated child seed, with separate train and evaluation streams derived beneath it. Scalar clean values are generated from the operational IID uniform stream,
\[
u\in[0,1],
\]
and mapped to
\[
x_0=2u-1.
\]
The resulting tensors have the same shape and value range as the corresponding real-data inputs. Probe training clean tensors are generated online from a continuous stream. Probe evaluation uses a separately seeded fixed bank of 4,096 clean tensors. The diffusion noise is generated independently from the tested source and enters through the standard forward process
\[
x_t
=
\sqrt{\bar\alpha_t}x_0
+
\sqrt{1-\bar\alpha_t}\epsilon_t.
\]
The model is trained with the same epsilon-prediction objective as in the real-data experiment.

The MNIST U-Net uses channel widths
\[
(64,128,128),
\]
with input shape
\[
1\times 28\times 28.
\]
The CIFAR-10 U-Net uses channel widths
\[
(128,256,256,256),
\]
with input shape
\[
3\times 32\times 32.
\]
All models are trained from scratch. Both datasets use a linear diffusion schedule with
\[
\beta_1=10^{-4},
\qquad
\beta_T=0.02,
\qquad
T=1000.
\]

Optimization uses AdamW with learning rate
\[
10^{-4},
\]
momentum parameters
\[
(0.95,0.999),
\]
weight decay
\[
10^{-6},
\]
a 500-step warm-up, cosine learning-rate decay, and gradient-norm clipping at 1.0. Mixed-precision training is used when CUDA is available.

MNIST is evaluated every 469 optimization steps and CIFAR-10 every 750 steps. Each evaluation reuses the same fixed clean examples, timesteps, and noise tensors for all checkpoints within a run. Tail loss is defined as the arithmetic mean of the final 10 recorded evaluation losses.

Standard generation produces 1,000 samples using 100-step DDIM sampling with
\[
\eta=0
\]
and IID initial noise. Experiments that replace the initial sampling source by the matched tested source are treated separately from the standard training protocol.

\subsection{Shuffle-ablation implementation}
\label{app:shuffle_protocol}

The shuffle ablation is applied after the source-specific marginal transform. For each tested diffusion-noise role, the complete scalar stream that would otherwise be consumed by that role is first materialized as a one-dimensional tape,
\[
(z_1,z_2,\ldots,z_N).
\]
An independent global permutation $\pi$ is then applied,
\[
(z_1,z_2,\ldots,z_N)
\longrightarrow
(z_{\pi(1)},z_{\pi(2)},\ldots,z_{\pi(N)}).
\]

The shuffled and original streams therefore contain exactly the same scalar values with exactly the same empirical marginal distribution. Only their positions and sequential dependence are changed. The permuted tape is subsequently consumed in the same order as an ordinary stream and reshaped into diffusion-noise tensors as required by the training or evaluation procedure.

Permutation is role-local. Training-noise and evaluation-noise tapes are shuffled independently, and values are never exchanged between distinct random roles. The permutation itself is generated by an independent PCG64DXSM stream with a role-specific child seed and does not use the tested source to determine the ordering.

For large tapes, the implementation uses a bounded-memory external permutation procedure, but this affects only memory management and not the experimental definition above.

\section{Finite-Precision Numerical Failure Cases}
\label{app:orbit_screening}

\subsection{Finite-precision absorbing states}

We exclude trajectories that collapse into absorbing states under finite-precision arithmetic. The numerical iteration is a map on machine-representable values rather than the exact real-valued dynamical system, and rounding can therefore create trajectories that do not occur in exact arithmetic \citep{ott2002,higham2002,li2005,kloewer2023}.

The risk is particularly large near a critical point $c$ satisfying
\[
f^{\prime}(c)=0.
\]
For a small displacement $h$,
\[
f(c+h)
=
f(c)
+
\frac{1}{2}f^{\prime\prime}(c)h^2
+
O(h^3).
\]
If $\Delta$ denotes the effective output-rounding tolerance, values satisfying
\[
|f(c+h)-f(c)|
\lesssim
\Delta
\]
can be mapped to the same machine-representable output. Neglecting higher-order terms gives the approximate capture width
\[
|h|
\lesssim
\sqrt{
\frac{2\Delta}{|f^{\prime\prime}(c)|}
}.
\]
By contrast, near a point with
\[
f^{\prime}(c)\neq 0,
\]
the corresponding width is approximately
\[
|h|
\lesssim
\frac{\Delta}{|f^{\prime}(c)|}.
\]
A critical point can therefore enlarge the rounding-induced capture region from order $\Delta$ to order $\sqrt{\Delta}$.

For the Sine-1 map,
\[
f(x)=\sin(\pi x)\bmod 1,
\]
the experiment reached
\[
x=0.5000000021116002.
\]
The floating-point evaluation of
\[
\sin(\pi x)
\]
rounded to exactly 1. The modulo operation then produced 0, after which the orbit remained permanently at the zero absorbing state.

For Logistic 4.0,
\[
f(x)=4x(1-x),
\]
the trajectory reached
\[
x=0.4999999944571299.
\]
The next floating-point evaluation rounded to exactly 1.0, after which the following iteration produced 0.0 and the orbit remained at zero.

These failures arise from the realized finite-precision trajectories rather than from the continuous comparison of source learnability studied in the main experiments. Sine-1 and Logistic 4.0 are therefore excluded from the main source set.

\section{Derivation of the IID Probe Limit}
\label{app:iid_probe_limit}

We derive the theoretical prediction limit used in Section~\ref{sec:mismatch}.
Consider one scalar component of the IID-clean diffusion probe at a fixed timestep $t$:
\begin{equation}
x = a_t u + b_t e,
\end{equation}
where
\begin{equation}
a_t = \sqrt{\bar{\alpha}_t},
\qquad
b_t = \sqrt{1-\bar{\alpha}_t},
\end{equation}
and
\begin{equation}
u \sim \mathrm{Uniform}(-1,1),
\qquad
e \sim \mathcal{N}(0,1).
\end{equation}
The variables $u$ and $e$ are independent. Since the coordinates are independent in this ideal reference problem, the full elementwise mean-squared loss reduces to the same scalar calculation.

For a fixed observed value $x$, the constraint
\begin{equation}
-1 \leq u \leq 1
\end{equation}
implies
\begin{equation}
\frac{x-a_t}{b_t}
\leq e \leq
\frac{x+a_t}{b_t}.
\end{equation}
Define
\begin{equation}
l_t(x)=\frac{x-a_t}{b_t},
\qquad
r_t(x)=\frac{x+a_t}{b_t}.
\end{equation}

Conditioned on $x$, the variable $e$ therefore follows a standard Gaussian truncated to the interval
\begin{equation}
[l_t(x),r_t(x)].
\end{equation}
Let $\phi$ and $\Phi$ denote the standard Gaussian density and cumulative distribution function. The conditional mean is
\begin{equation}
m_t(x)
=
\mathbb{E}[e\mid x,t]
=
\frac{
\phi(l_t(x))-\phi(r_t(x))
}{
\Phi(r_t(x))-\Phi(l_t(x))
}.
\end{equation}

The density of $x$ is obtained by integrating over the admissible values of $e$:
\begin{equation}
p_t(x)
=
\frac{
\Phi(r_t(x))-\Phi(l_t(x))
}{
2a_t
}.
\end{equation}

Under squared-error loss, the Bayes-optimal predictor is the conditional mean $m_t(x)$. Therefore the minimum prediction loss at timestep $t$ is
\begin{equation}
L_t^{*}
=
\mathbb{E}
\left[
\left(
e-\mathbb{E}[e\mid x,t]
\right)^2
\right].
\end{equation}
Using
\begin{equation}
\mathbb{E}[e^2]=1,
\end{equation}
and the orthogonality property of conditional expectation,
\begin{equation}
L_t^{*}
=
1-
\mathbb{E}
\left[
m_t(x)^2
\right].
\end{equation}
Hence
\begin{equation}
L_t^{*}
=
1-
\int_{-\infty}^{\infty}
m_t(x)^2 p_t(x)\,dx.
\end{equation}

The diffusion timestep is sampled uniformly from
\begin{equation}
t\in\{0,\ldots,T-1\},
\end{equation}
so the theoretical IID probe limit is
\begin{equation}
L_0
=
\frac{1}{T}
\sum_{t=0}^{T-1}
L_t^{*}.
\end{equation}

For the linear $T=1000$ diffusion schedule used in our experiments, with
\begin{equation}
\beta_1=10^{-4},
\qquad
\beta_T=0.02,
\end{equation}
numerical integration gives
\begin{equation}
L_0 \approx 0.184016.
\end{equation}

This value is the Bayes-optimal population prediction limit for the ideal IID-clean and IID-Gaussian probe. It is not the initial loss of an untrained network.

\section{Modified-Objective Ablation and CDM Loss Adaptation}
\label{app:cdm}
\begin{figure}[!ht]
    \centering
    \begin{subfigure}[t]{0.19\linewidth}
        \centering
        \captionsetup{font=scriptsize}
        \includegraphics[width=\linewidth]{figures/5/mr_gr_1_off_grid.png}
        \caption{GR, standard}
    \end{subfigure}
    \hfill
    \begin{subfigure}[t]{0.19\linewidth}
        \centering
        \captionsetup{font=scriptsize}
        \includegraphics[width=\linewidth]{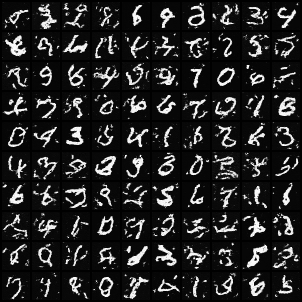}
        \caption{GR, modified}
    \end{subfigure}
    \hfill
    \begin{subfigure}[t]{0.19\linewidth}
        \centering
        \captionsetup{font=scriptsize}
        \includegraphics[width=\linewidth]{figures/5/mr_m1_1_off_grid.png}
        \caption{M1, standard}
    \end{subfigure}
    \hfill
    \begin{subfigure}[t]{0.19\linewidth}
        \centering
        \captionsetup{font=scriptsize}
        \includegraphics[width=\linewidth]{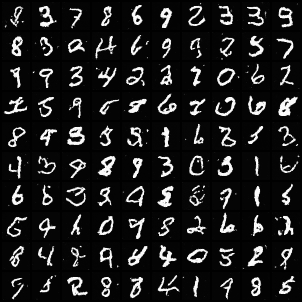}
        \caption{M1, modified}
    \end{subfigure}
    \hfill
    \begin{subfigure}[t]{0.19\linewidth}
        \centering
        \captionsetup{font=scriptsize}
        \includegraphics[width=\linewidth]{figures/5/mr_iid_1_off_grid.png}
        \caption{IID baseline}
    \end{subfigure}
    \caption{Modified-objective ablation with IID initial noise throughout. Panels (a,b) compare standard and modified objectives for GR; (c,d) give the corresponding M1 comparison. The modified objective is defined in Equation~\ref{eq:modified_loss}. Panel (e) shows the standard-objective IID-trained baseline. These representative MNIST examples illustrate the dependence of generation behavior on the training objective.}
    \label{fig:modified_objective_examples}
\end{figure}

This ablation examines whether changing the objective can alter the
generation degradation observed under IID sampling. It retains the
standard noise-prediction term while adding clean-prediction consistency
regularization and a two-sample clean-target term. The latter uses the
true training image $x_0$ to penalize the squared bias of the endpoint
conditional mean, rather than the ordinary mean-squared error of each
individual endpoint. Figure~\ref{fig:modified_objective_examples} shows improved
representative outputs for GR and M1. Because the two additional terms
are introduced together, the comparison does not identify their separate
contributions or quantify a division between learning clean-data
regularities and exploiting noise structure. It serves as a supporting
sensitivity check rather than a proposed replacement objective.

The modified-loss experiment uses a discrete VP adaptation of the consistency regularization in Consistent Diffusion Models~\citep{daras2023consistent}.
For a noisy input $x_t$, the model first predicts $\epsilon_\theta(x_t,t)$ and obtains the corresponding clean estimate
\begin{equation}
h_0
=
\frac{
x_t-\sqrt{1-\bar{\alpha}_t}\,
\epsilon_\theta(x_t,t)
}{
\sqrt{\bar{\alpha}_t}
}.
\end{equation}

From the same $x_t$, two independent stochastic reverse trajectories are then generated to an earlier timestep $s<t$.
Let $h_1$ and $h_2$ denote the clean estimates obtained at the two trajectory endpoints.
The consistency term is
\begin{equation}
\mathcal{L}_{\mathrm{CDM}}
=
\frac{1}{d}
\sum_{j=1}^{d}
(h_{1,j}-h_{0,j})
(h_{2,j}-h_{0,j}).
\end{equation}

The two trajectories are reused for the additional clean-image term.
Using the unclipped endpoint clean estimates $\widetilde h_1$ and $\widetilde h_2$, we define
\begin{equation}
\mathcal{L}_{x_0}
=
\frac{1}{2d}
\sum_{j=1}^{d}
(\widetilde h_{1,j}-x_{0,j})
(\widetilde h_{2,j}-x_{0,j}).
\end{equation}

Because the two endpoint trajectories are conditionally independent,
this cross-product is a two-sample estimator of the squared deviation
of the endpoint conditional mean from the clean target $x_0$, up to
the factor $1/2$. Unlike an ordinary endpoint MSE, it does not directly
penalize the conditional variance of the endpoint predictions. The complete training objective is
\begin{equation}
\mathcal{L}
=
\mathcal{L}_{\epsilon}
+
2\mathcal{L}_{\mathrm{CDM}}
+
0.1\mathcal{L}_{x_0},
\label{eq:modified_loss}
\end{equation}
where $\mathcal{L}_{\epsilon}$ is the standard noise-prediction MSE.

In the experiment, the local reverse rollout uses six discrete points, a maximum timestep drop of $50$, and stochastic DDIM transitions with $\eta=1$.
The original CDM consistency term uses clipped clean estimates in the range $[-1,1]$, while the additional $x_0$ term uses the corresponding unclipped endpoint estimates.

\end{document}